\documentclass[letterpaper, 10 pt, conference]{IEEEtran}

\usepackage{amsmath} 
\usepackage{amssymb}  
\usepackage[section]{placeins}
\usepackage[pdftex]{graphicx}
\usepackage{todonotes}
\usepackage{url}
\usepackage{subfig}
\usepackage[para]{footmisc} 
\graphicspath{{Plots/}{Figures/}}
\DeclareGraphicsExtensions{.pdf,.jpeg,.png,.PNG,.jpg}

\title{\LARGE \bf
A Robot Localization Framework Using CNNs for Object Detection and Pose Estimation
}

\author{\IEEEauthorblockN{Lukas Hoyer}
\IEEEauthorblockA{\textit{Faculty of Computer Science} \\
\textit{Otto von Guericke University}\\
Magdeburg, Germany \\
lukas.hoyer@st.ovgu.de}
\and
\IEEEauthorblockN{Christoph Steup}
\IEEEauthorblockA{\textit{Faculty of Computer Science} \\
\textit{Otto von Guericke University}\\
Magdeburg, Germany \\
steup@ovgu.de}
\and
\IEEEauthorblockN{Sanaz Mostaghim}
\IEEEauthorblockA{\textit{Faculty of Computer Science} \\
\textit{Otto von Guericke University}\\
Magdeburg, Germany \\
sanaz.mostaghim@ovgu.de}
}

\begin{document}

\maketitle
\thispagestyle{empty}
\pagestyle{empty}

\begin{abstract}

External localization is an essential part for the indoor operation of small or cost-efficient robots, as they are used, for example, in swarm robotics. We introduce a two-stage localization and instance identification framework for arbitrary robots based on convolutional neural networks. Object detection is performed on an external camera image of the operation zone providing robot bounding boxes for an identification and orientation estimation convolutional neural network. Additionally, we propose a process to generate the necessary training data. The framework was evaluated with 3 different robot types and various identification patterns. We have analyzed the main framework hyperparameters providing recommendations for the framework operation settings. We achieved up to 98\% mAP@IOU0.5 and only 1.6$^\circ$ orientation error, running with a frame rate of 50~Hz on a GPU.

\end{abstract}


\section{INTRODUCTION}

Indoor robotics represents a field with special properties and challenges
regarding locomotion. In contrast to outdoor robotics, where the unknown
structure and dynamics of the environment are the largest problems for moving a
robot from point $A$ to point $B$, indoor robotics mainly needs to overcome the lack
of reliable position estimation. This is caused by the lack of
GPS reception as well as the usually smaller error margins enforced by
the smaller size of the working area. However, position alone is not
enough to establish reliable locomotion. Robots also need an estimate of their
orientation to enable predictable movements. The composition of these two
parameters is called pose and is the basis of all motion planning.

This paper introduces a deep neural network based pose estimation system for
swarms of robots, enabling the detection of position, orientation as well as type
and instance identification of each robot. It utilizes cameras mounted in
the operation environment of the robots. External localization is particularly suitable
for our scenario as swarm robots are usually too small or
too cost-efficiently built to provide sophisticated sensors for self-localization.
Even though multiple camera-based indoor
pose estimation systems already exist (see Section~\ref{sota}), the proposed
pose estimation framework provides some special characteristics such as easy adaptivity
to different types of robots, identification properties and environments as well as
fast processing speed to provide update rates and latencies similar to fast
GPS devices. The versatility towards identification markers enables the system
to handle a wide range of properties to separate robots of the same type such as:
colors, numbers, letters and LED patterns. Additionally, we provide a process description to straightforwardly generate the
necessary training data. It includes automation and augmentation
steps, which are evaluated regarding their performance for a test system of
three different robot types with three different marker types in an actual
robotics lab. Especially, the ability to easily change the used type of marker
attached to the robot is beneficial to real-world applications, as some
marker types may not be usable in certain scenarios, because of occlusion,
illumination or size of the robot.

The rest of the paper is structured as follows. Section~\ref{sota}
describes relevant existing approaches in the field of external camera-based
localization and pose estimation. Section~\ref{framework} presents the details
of the proposed framework and the training and adaptation process. The paper
ends with an evaluation of different configurations of the framework and of
different process steps to acquire training data in Section~\ref{eval} as well as a
conclusion including future work in Section~\ref{conclusion}.

\section{RELATED WORK}
\label{sota}

\begin{figure*}
\centering
\includegraphics[width=6.5in]{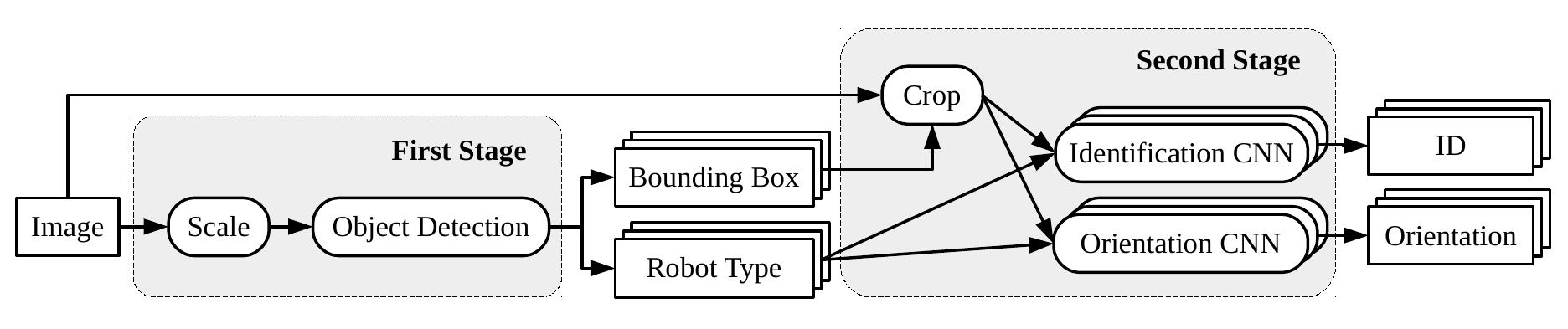}
\caption{Architecture of the robot localization framework consisting of two stages. The first stage samples down the camera frame and looks for robots. The detection results are processed by the second stage. For each robot, the frame is cropped according to the bounding box and fed into a robot type specific identification and an orientation estimation CNN.}
\label{fig_architecture}
\vspace{-0.4cm}
\end{figure*}

Currently, explicit systems for robot localization exist, which provide high
update rates and high precision, such as
WhyCon~Krajnik~et~al.~\cite{krajnik2014practical} and Vicon.
However, those do not provide flexibility in choosing the used identification
marker for the robots. Depending on the type of robot used, it may be
challenging or even impossible to attach the necessary markers.
Therefore, a flexible solution, handling arbitrary identification properties, is
beneficial for robotic labs.

Convolutional neural networks (CNN) may provide a viable alternative if the pose
of the robot can be detected. To this end, pose estimation and its counterpart,
viewport estimation, were studied in the deep neural network community. The
research showed two different concepts, the separate and the simultaneous
approach, which differ in the integration of the pose estimation step.
Glasner~et~al.~\cite{glasner2012viewpoint},
Tulsiani~and~Malik~\cite{tulsiani2015viewpoints},
Tulsiani,~Carreira~and~Malik~\cite{tulsiani2015pose},
Redondo-Cabrera~and~L\'opez-Sastre~\cite{redondo2015boosted} and Poirson~et~al.~\cite{poirson2016single} provide examples of the separated
approach, where the pose estimation step is executed separately after the
object detection and classification is done. The approaches of
Pepik~et~al.~\cite{pepik2012teaching},
Xiang,~Mottaghi~and~Savarese~\cite{xiang2014beyond},
Massa,~Marlet~and~Aubry~\cite{massa2016crafting} and
Xiang~et~al.~\cite{xiang2018posecnn} are examples of the simultaneous approach,
which embeds the pose estimation within the object detection and classification
network. O\~noro-Rubio~et~al.~\cite{onoro2018challenge} created a set of three
approaches that represent similar systems following the separated, the
simultaneous and a mixed strategy, working out that a model with separated detection and orientation estimation achieves a higher overall performance. Additionally, some special solutions were created.
Su~et~al.~\cite{su2015render} proposed a system that only uses synthetic data to
train the network and provided comparable results to systems trained on
real-world data. \emph{SSD-6D} of Kehl~et~al.~\cite{kehl2017ssd6d} is a special
system that is able to integrate depth information in addition to
pure RGB data.

Most of the listed approaches are based on previous work about object detection using CNNs. While \cite{kehl2017ssd6d}, \cite{poirson2016single} and \cite{tekin2017realtime} utilize a single shot architecture as outlined in \cite{liu2016ssd}, the approaches described in \cite{massa2016crafting}, \cite{onoro2018challenge} and \cite{su2015render} are built upon R-CNN \cite{girshick2014rich} and its successors.

Our approach uses a separated deep neural network approach, but with the aim
of minimizing the effort for training data generation and to reduce latency to
build a system customized for robotics. These goals are not the focus of the
present approaches. All existing concepts use generic benchmark datasets such as
Pascal VOC \cite{everingham2010pascal} and MS~COCO \cite{lin2014microsoft} to test
the algorithms. Therefore, they
did not care to minimize the effort for training data generation. Our
approach, however, particularly considers the whole process from training data
acquisition to training the two-stage network for detection and pose
estimation.

\section{PROPOSED LOCALIZATION FRAMEWORK}
\label{framework}

\subsection{Architecture}
\label{architecture}

The localization framework is based on fixed RGB cameras looking at the robot operation zone. On the recorded camera frames, we perform a two-stage image processing, consisting of an object detection localizing the robots within the scene (first stage) and an instance classification as well as an orientation estimation (second stage) as shown in figure \ref{fig_architecture}.

The first stage performs a low-resolution robot detection. It samples down the camera image and uses a CNN object detection to estimate the type (e.g. quadcopter or wheeled robot) and the bounding box of a robot. The results of the first stage are used to feed the second stage. The original, high-resolution camera image is cropped according to the estimated bounding boxes. Depending on the predicted robot type, a corresponding second stage neural network is selected. For each robot type, there are a single instance identification and orientation estimation CNN. In the end, the output of both stages is merged resulting in a bounding box, the type, the instance identification and the orientation of each robot.

The orientation estimation neural network supports, both, continuous and discrete pose estimation. To provide a continuous orientation, a linear output activation function and the mean squared error of the smallest angle difference are used.
For the discrete approach, the orientation angles are separated into 360 bins resulting in a classification problem. The performance of both approaches is compared in Section \ref{eval}.

As the cameras are fixed, their position relative to the ground plane can be determined. For robots moving on the floor, this information is sufficient to calculate the 3D position and the projected 2D orientation. For flying robots such as quadcopter, their height can be provided by an on-board sensor. In the experiments, only one camera is used, but multiple cameras can be supported by merging the output of the framework after localization.

We have decided to use the two-stage approach, as it provides several advantages over an integrated one-stage approach. Normally, robots represent only a small part of the camera frame. Therefore, most of the pixels do not contain relevant information. As comparably low-resolution features are sufficient to recognize the robot itself, the first stage robot detection provides a fast way to find relevant image sections. This information can be utilized to provide high-resolution image crops for identification and orientation estimation of a detected robot, which depend on smaller features than the detection of the whole robot. The speed increase due to the first stage screening justifies the recalculation of convolutional layers with a higher resolution in the second stage. Moreover, this approach allows that the neural networks of first and second stage can be developed and optimized independently. Furthermore, due to the separation of identification and orientation estimation for each robot type, the CNNs are more specialized and can be chosen with less capacity as they handle less features and classes.

To provide a fast inference speed, we use state-of-the-art lightweight CNNs. The first stage is based on SSD \cite{liu2016ssd} with MobileNet v2 \cite{sandler2018mobilenetv2} as feature extractor. This decision is based on the comparison of different object detection meta-architectures and their feature extractors in \cite{huang2017speed}. The chosen architecture/model is the fastest configuration in the Pareto front. For the second stage, a MobileNet is used for the same reasons. Moreover, the width of MobileNet can be easily adjusted by one parameter, allowing speed/performance adjustments \cite{howard2017mobilenets}. We have used the TensorFlow Object Detection API for the implementation of the first stage and Keras to realize the second stage as they provide implementations of the used network architectures and pretrained network weights.

\subsection{Training Data Acquisition Process}

A crucial contribution to the performance of a neural network is the amount and quality of its training data \cite{sun2017revisiting}. Usually, labeling enough images to train a CNN requires a lot of effort, which is not reasonable for setting up a robot localization framework. To avoid this problem, we synthesize training data by superimposing image crops, containing robots, on background images. The necessary robot crops are extracted from images of robots in the working area, which are recorded with the localization camera during a setup phase.

\subsubsection{Compositor}

The compositor is an algorithm that superimposes robot crops on background images with random scale, rotation and position. It is similar to the approach described in \cite{georgakis2017synthesizing}, but with context-insensitive, random placement. The compositor can generate a large amount of data with multiple robots in one frame and arbitrary robot position, orientation, size and occlusion. Its main advantage is the known ground truth information of the synthesized data, which saves a lot of time for labeling the images.

There are other methods to synthesize data such as rendering training images, discussed in \cite{peng2015learning} or \cite{richter2016playing}. But they have the disadvantage that it is quite difficult to find parameters to match the lighting in the robot working zone and the camera properties. Moreover, they require 3D models of all robots and material models for photo-realistic rendering. Therefore, it is easier to use real images of robots and to crop them.

Even though the amount of generated data is theoretically unlimited, the diversity of the synthesized data is restricted by the number of backgrounds and robot crops. Moreover, the robot crop does not necessarily match with the background like in a real image. This can affect lighting, blur/blending and crop/compositing artifacts as well as contextual information, which could influence the performance of the CNNs \cite{georgakis2017synthesizing}. For example, an occluded robot could be detected by its shadow that isn't generated by the compositor. However, these are only minor problems that are worthwhile to minimize the effort of labeling real images.

In order to minimize the bias induced by the class imbalance of the training data \cite{batista2004study}, the compositor balances all important factors for the experiments such as background types, robot types and identification patterns within one robot type. For each generated frame, a random background is chosen on which one to four randomly selected robot crops are composited with random scale, orientation and translation. The bounds of the robot size depend on possible distances of the robot to the camera. For each robot in a frame its type, identification, bounding box and orientation are stored as ground truth for the training.

\subsubsection{Robot Crops}

Robot crops are the essential factor for the quality and the necessary setup effort generating the final composited data. To extract them, a manual or an automatic approach may be used.

For the manual robot crop, an image sequence of a non-moving robot with a static or a changing identification pattern (e.g. realized with LEDs) is captured. For each image sequence, one single instance mask is manually created using an image editing tool. Later, this mask is applied to all images of the sequence.

The automatic robot crop is based on background subtraction. Firstly, an image of the background without robot is captured. After that, the robot is placed in the scene and an image sequence is recorded. Finally, the background image is subtracted from the image sequence followed by thresholding and morphological operations to generate an instance mask for each frame. Although, using the automatic method, robots could even move during the image sequence, only non-moving robots were captured due to comparability between both techniques.

To properly superimpose the robot crops on arbitrary backgrounds, the crop itself should not contain any background of the operation zone. Therefore, in both methods, the generated masks are used as alpha channel for each image in the sequence with the robot. The image is cropped using the bounding box of the outer contour of the robot mask. It has to be ensured that the robot crops are aligned in the same direction. This can either be done before the recording or by rotating the crops later. We recommend aligning the robot in scene as it saves some time.
%
%
\section{EVALUATION}
\label{eval}

\subsection{Evaluation Environment}

\begin{figure}[t!]
\centering
\includegraphics[width=0.7in]{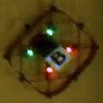}
\includegraphics[width=0.7in]{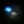}
\includegraphics[width=0.7in]{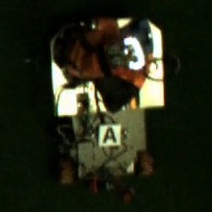}

\noindent
\includegraphics[width=0.7in]{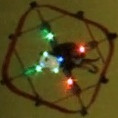}
\includegraphics[width=0.7in]{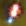}
\includegraphics[width=0.7in]{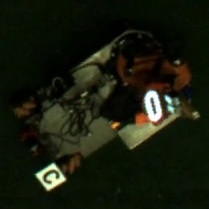}
\caption{Examples of each robot type with different identification patterns. From left to right: quadcopter (marker B and LED pattern blue-green), Sphero (LED black and LED bright red) and YouBot (maker A and marker C). Notice that the blue LED at the Spheros marks their rear and the marker is attached at different position at the YouBot.}
\label{fig_robots}
\end{figure}

We used three different robot types in our experiments: quadcopters, Spheros 
and a Kuka YouBot 
(see Figure \ref{fig_robots}). The quadcopters have 4 position LEDs (green and red) next to their rotors as well as four RGB identification LEDs in the center. The Spheros are spherical robots driven by a weight moving inside their transparent hull. They provide one blue position LED in the back and one RGB identification LED in the center. The YouBot is a wheeled robot with a manipulation arm in the front. The robot operation zone was recorded by a Basler acA1600-60gc standard industry camera, which was mounted on the ceiling.

As identification properties for our experiments, we used 15 different LED patterns as well as three letters (A,B,C) for the quadcopter, eight different LED colors for the Sphero and three letters for the YouBot. The identification markers of the YouBot were attached either in the center or on top of a pole at the back. Some example patterns are shown in Figure \ref{fig_robots}.

We have chosen these robot types and identification patterns as they provide versatile challenges for the localization framework to prove that it can be applied to a high variety of robots. The Spheros are quite small (about 25x25 pixels in the camera image) and don't offer the possibility to attach an identification marker due to their shape and locomotion. The quadcopters have a medium, but varying size (depending on their altitude) and provide many identification patterns, which are in some cases difficult to distinguish due to overexposure, reflections, spatial separation of the single LED color channels, blending with the position LEDs and occlusion. Finally, the YouBot provides an example for a bigger robot (about 200x120 pixels) with many visual features. It has only static markers as the standard YouBot does not provide RGB LEDs.

\subsection{Training and Evaluation Dataset}

For the training process, we have captured robot crops under different lighting conditions (artificial lighting and natural lighting), different locations of the robot in the operation zone (each corner and in the middle) and different floor colors (quadcopter yellow, Sphero yellow and green, YouBot green). For the compositor, 1524 images of the operation zone and 25,608 randomly sampled images from the 2017 training dataset of MS~COCO \cite{lin2014microsoft} were used as backgrounds. Which components should be used and how many training images should be generated, is analyzed in the experiments (see Section \ref{results}).

Depending on the stage of the framework, the training images are further processed. The first stage exploits random horizontal and vertical flips and SSD random crop \cite{liu2016ssd} to augment the dataset. The data for the second stage is cropped using the ground truth boxes with a variance from -10\% (inwards) to +15\% (outwards) and prepared according to the second stage pipeline (see Section \ref{architecture}).

The evaluation dataset for the first stage consists of 1400 images per robot type. The second stage evaluation set contains 110 images per identification pattern of each robot. Within both datasets, there is the same amount of frames with natural/artificial lighting, floor colors (Sphero) and pattern position (YouBot) for each robot type. The images of the evaluation set were semi-automatically labeled and manually corrected.

As the normal use case contains few objects in the robot working zone, additional objects were added to mislead the first stage and evaluate its specificity. We have chosen to use synthetic objects, which were extracted from the eval2017 dataset of MS~COCO \cite{lin2014microsoft}, as they provide a good variety of decoy objects. Three randomly chosen decoys were superimposed at a random position and rotation on each evaluation image while it was ensured that they do not occlude the robot.

\subsection{Experiment Setup}
\label{exp_setup}

We have chosen the experiments, presented in this section, to provide recommendations for a good configuration of the localization framework. Therefore, we have evaluated the main parameters of the compositor as well as important hyperparameters of the two framework stages.
\begin{table}
\caption{Default experiment configuration}
\label{table_default_conf}
\begin{center}
\begin{tabular}{|l||c|c|}
\hline
 & First Stage & Second Stage\\
\hline \hline
Batch size & 16 & 32\\
\hline
Optimizer & rmsprop & adam\\
\hline
Learning rate & 0.004/0.0004 (after 15k steps) & 0.0004\\
\hline
Input size & 400x300 & 128x128\\
\hline
Training data & 5k images & 2k crops per id\\
\hline
Repetitions & 5 & 8\\
\hline
Evaluation period & step 20k - 25k & epoch 15 - 25\\
\hline
\end{tabular}
\end{center}
\end{table}
For both stages, a default/reference configuration was chosen as base level. It is annotated with F/S~0. In all experiment abbreviations, F means first stage and S stands for the second stage. Table \ref{table_default_conf} lists all important default settings and Table \ref{table_data_exp} illustrates which parameters were changed in each experiment configuration.

The first three experiment configurations (F/S~1 -- F/S~3) analyzed the behavior of both stages with respect to the compositor settings. In F/S~1, the influence of the background diversity of the generated images was researched. Therefore, training data generated with backgrounds only of the robot working zone (SwarmLab), only of MS~COCO and both combined was studied. We wanted to figure out whether a background set with higher diversity (MS~COCO) can improve the learning process and if the framework has to be trained with scenario-specific backgrounds. F/S~2 examined the effect of the amount of composited images to analyze the number of useful images when their impact on the performance is limited by the diversity of crops and backgrounds. In the third experiment (F/S~3), the influence of the crop method was surveyed determining whether a more precise mask generated by manual labeling justifies its additional cost or if the more versatile automatic crops can even improve the performance of the framework.

The subsequent experiments were conducted to analyze hyperparameters of the framework architecture. In F~4, the input resolution of the first stage was studied to find a good speed-accuracy trade-off. For the same reason, in S~4, the influence of the MobileNet width multiplier $\alpha$ \cite{howard2017mobilenets} of the second stage was researched. Finally, S~5 evaluated the performance of continuous and discrete pose estimation.

For the first stage, SSD with MobileNet~v2, which was pre-trained on MS~COCO from \cite{huang2017speed}, was used. To adapt the pre-trained network to our problem, the number of neurons of the classification output layer was adjusted to match the number of robot types and the entire network was fine-tuned with a low learning rate. The chosen hyperparameters for the training can be seen in Table \ref{table_default_conf}. After 20,000 training steps (one batch iteration) the network wasn't improving anymore. The evaluation data for the experiments was extracted from step 20,000 to step 25,000 every 250 steps.

The second stage is based on network weights that were pre-trained on ImageNet provided by Keras. As the network wasn't improving after 15 dataset iterations (epochs), the evaluation was based on epoch 15 to 25. The MobileNet input size is 128 because bigger robot crops do not provide more necessary details as the robots itself are usually smaller than 200x200 pixels.

The first stage detection performance was evaluated with Pascal VOC 2010 mAP@0.5IOU \cite{everingham2010pascal} (mean average precision for an intersection over union of at least 0.5) and for the second stage the classification accuracy as well as the mean absolute error of the smallest angle difference were used.
\begin{table}
\caption{Experiment overview}
\label{table_data_exp}
\begin{center}
\begin{tabular}{|c||l|}
\hline
Experiment & Description\\
\hline \hline
F/S~0 & SwarmLab and COCO compositor backgrounds\\
F/S~1.1 & only SwarmLab compositor backgrounds\\
F/S~1.2 & only COCO compositor backgrounds\\
\hline
F/S~2.1 & 1/5 of the default image amount\\
F 0 & default image amount (see Table \ref{table_default_conf})\\
F/S~2.2 & 4 times of the default image amount\\
\hline
F/S~0 & manual crop\\
F/S~3.1 & automatic crop\\
F/S~3.2 & automatic and manual crop\\
\hline
F 4.1 & first stage resolution of 200x150 pixels\\
F 0 & first stage resolution of 400x300 pixels\\
F 4.2 & first stage resolution of 800x600 pixels\\
\hline
S~4.1 & MobileNet width multiplier $\alpha = 0.25$\\
S~0 & MobileNet width multiplier $\alpha = 0.5$\\
S~4.2 & MobileNet width multiplier $\alpha = 0.75$\\
S~4.3 & MobileNet width multiplier $\alpha = 1.0$\\
\hline
S~0 & discrete orientation estimation (classification)\\
S~5 & continuous orientation estimation (regression)\\
\hline
\end{tabular}
\end{center}
\end{table}

\subsection{Results}
\label{results}

\subsubsection{First Stage}
\label{first_stage}

\begin{figure}
\centering
\begin{minipage}{\columnwidth}
  \subfloat[First stage detection mean average precision for quadcopters.]{%
    \label{fig_fstage_copter}%
    \includegraphics[width=3.2in]{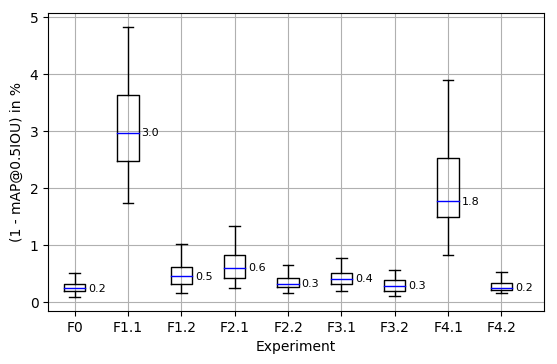}}\\
  \subfloat[First stage detection mean average precision for Spheros.]{%
    \label{fig_fstage_sphero}%
    \includegraphics[width=3.2in]{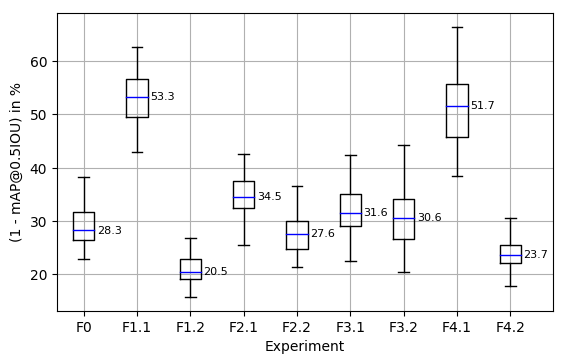}}\\
  \subfloat[First stage detection mean average precision for YouBots.]{%
    \label{fig_fstage_youbot}%
    \includegraphics[width=3.2in]{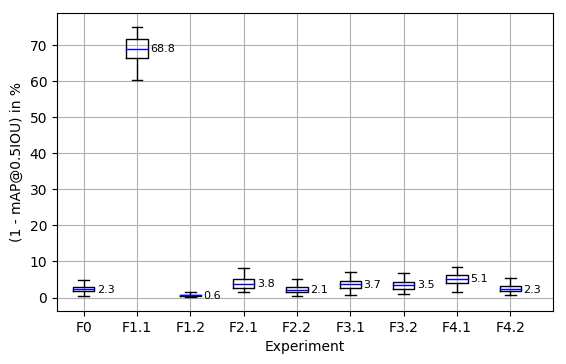}}\\
\end{minipage}
\caption{Performance comparison of the first stage.}
\label{fig_first_stage}
\end{figure}

In this section, the performance of the first stage is evaluated. With the best configuration, it reaches a median mAP@0.5IOU of 99.8\% (copter) / 79.5\% (Sphero) / 99.4\% (YouBot). Compared to the other robot types, the detection rate of the Sphero is quite low. This has several reasons. First of all, the Spheros are considerably smaller, which means that more precise bounding boxes in relation to the image size are necessary to achieve an intersection over union of at least 50\%. Moreover, the Spheros have only few features and, in addition, there are some objects containing similar features e.g. LEDs, light bulbs or light reflections. So it is a quite challenging task to detect the Spheros among other objects.

In the experiments, the first stage performance with respect to different configurations was compared as can be seen in Figure \ref{fig_first_stage}. One important factor is the background set for the compositor. It was figured out that only MS~COCO backgrounds (F1.2) work better (lower median and less variance) than only SwarmLab backgrounds (F1.1), as the SwarmLab backgrounds do not contain other decoys while COCO contains a variety of different objects resulting in a better generalization. Another important factor is the number of composited images (F2.1 / F0 / F2.2). Generally, the performance of the first stage was increased by using more images as it is also suggested in \cite{sun2017revisiting}.

As crop method, manual cropping (F0) was figured out to achieve a lower median and variance in comparison to automatic crop (F3.1) and both (F3.2). Nevertheless, the difference is tolerable and, therefore, it is reasonable to use automatic crops to speed up the setup of the framework.

In the last experiment for the first stage, different input resolutions were compared. The presumption that larger robots need less resolution for detection was confirmed. Even with 200x150 pixels (F4.1), the YouBot achieved good results while 400x300 pixels (F0) provided the best mAP@0.5IOU for copters. Small robots such as the Sphero need a high resolution of 800x600 pixels (F4.2) to provide the best performance, as more small features can be exploited by the CNN. However, the input resolution substantially affects the inference speed of the first stage as can be seen in Table \ref{table_benchmark_first_stage}. Therefore, we have chosen 400x300 pixels as default configuration to provide the best speed/accuracy trade-off for the first stage. In that case, the first stage takes 12~ms on a Nvidia GeForce GTX 1080 or 76~ms on an Intel Xeon E3-1230~v3~@~3.30GHz.

All in all, we recommend to generate about 20,000 images (depending on scenario, number of crops and backgrounds), use a diverse background set (no specific backgrounds of robot working zone are necessary), automatic crop and an input resolution of 400x300 pixels for the first stage.

\subsubsection{Second Stage}
\label{second_stage}

\begin{table}
\caption{Benchmark of the first stage}
\label{table_benchmark_first_stage}
\begin{center}
\begin{tabular}{|l||c|c|c|}
\hline
First stage resolution & 200x150 & 400x300 & 800x600\\
\hline \hline
Runtime on Xeon E3-1230 v3 (ms) & 27.36 & 76.35 & 322.11\\
\hline
Runtime on GTX 1080 (ms) & 10.27 & 12.26 & 22.89\\
\hline
\end{tabular}
%
%
\end{center}
\end{table}

\begin{figure*}
  \centering
  \begin{minipage}{\columnwidth}
  \subfloat[Second stage identification performance for quadcopters.]{%
    \label{fig_second_stage_copter_cat}%
    \includegraphics[width=3.4in]{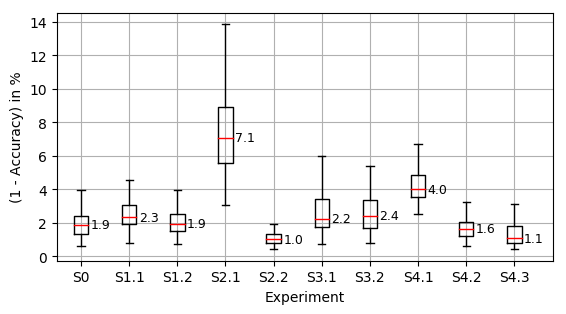}}\\
  \subfloat[Second stage identification performance for Spheros.]{%
    \label{fig_second_stage_sphero_cat}%
    \includegraphics[width=3.4in]{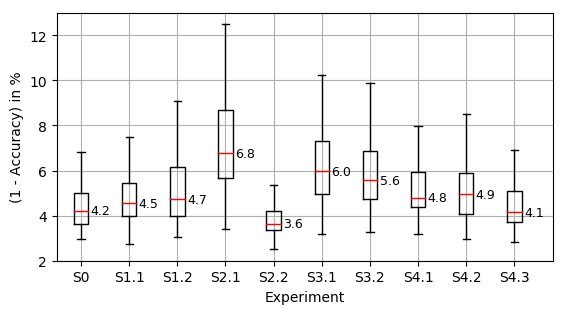}}\\
  \subfloat[Second stage identification performance for YouBots.]{%
    \label{fig_second_stage_youbot_cat}%
    \includegraphics[width=3.4in]{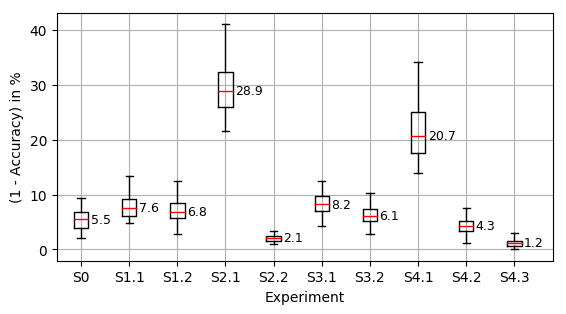}}
  \end{minipage}
  \begin{minipage}{\columnwidth}
  \subfloat[Second stage orientation estimation performance for quadcopters.]{%
    \label{fig_second_stage_copter_bin}%
    \includegraphics[width=3.4in]{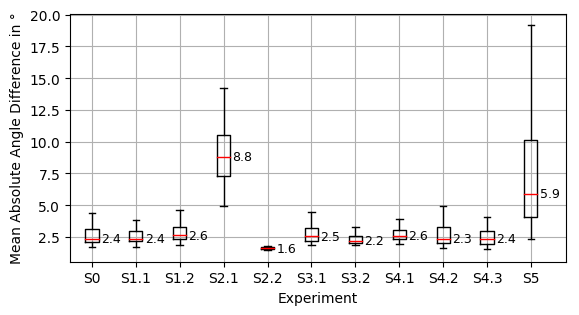}}\\
  \subfloat[Second stage orientation estimation performance for Spheros.]{%
    \label{fig_second_stage_sphero_bin}%
    \includegraphics[width=3.4in]{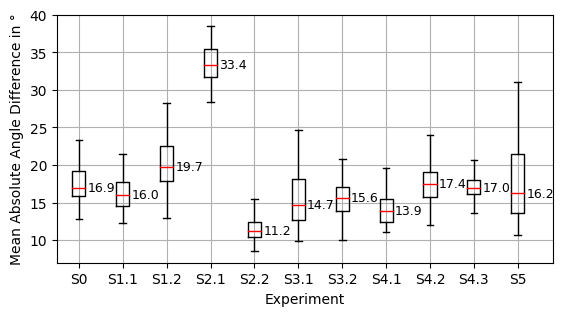}}\\
  \subfloat[Second stage orientation estimation performance for YouBots.]{%
    \label{fig_second_stage_youbot_bin}%
    \includegraphics[width=3.4in]{second_stage_copter_bin}}
  \end{minipage}
  \caption{Performance comparison of the second stage for identification and orientation estimation with respect to Table \ref{table_data_exp}.}
  \label{fig_second_stage}
\vspace{-0.5cm}
\end{figure*}

The second stage achieves good results for its instance identification and orientation estimation. With the best configuration, it reaches a median instance identification accuracy of 98.9\% (copter) / 96.4\% (Sphero) / 98.8\% (YouBot) and a mean absolute orientation error of 1.6$^\circ$ (copter) / 11.2$^\circ$ (Sphero) / 2.3$^\circ$ (YouBot).

The comparably low instance identification and orientation estimation results of the Sphero are probably caused by its locomotion as the LED board is sometimes lopsided and the identification LED is not properly visible. The same is true for the position LED, which complicates the orientation estimation. Moreover, the position LED is in some cases outshone by a bright identification LED color.

In the experiments, it was found out that a higher amount of composited images (S2.1 / S0 / S2.2) significantly improves the instance identification accuracy and the orientation estimation for all robot types as shown in Figure \ref{fig_second_stage}. The number of images is the most important factor influencing the performance of the framework decreasing both the median and the variance of the error. It should be noted that even the lowest amount of generated images exceeds the number of crops by far.

The second most important factor is the network size researched in S4.1 / S0 / S4.2 / S4.3. A network with more feature maps improves the instance identification performance especially for the YouBot a lot while it has less influence on Spheros (see Figure \ref{fig_second_stage_copter_cat} - \ref{fig_second_stage_youbot_cat}). This could be caused by the amount of visual features of the different robot types. There are more features necessary to classify the letter markers of the YouBot than the single RGB identification LED of the Sphero. Therefore, more features must be learned by the neural network, which requires a broader architecture with more feature maps. For orientation estimation, a bigger network for Spheros even deteriorates the performance (see Figure \ref{fig_second_stage_sphero_bin}), probably as there are less features necessary for the orientation estimation and larger networks tend to overfit the problem.

The influence of the background and the crop method are less substantial. Using all available backgrounds (S0 vs S1.1 and S1.2) improves the instance identification performance slightly (see Figure \ref{fig_second_stage_copter_cat} - \ref{fig_second_stage_youbot_cat}) as it adds variance to the training dataset. For identification purpose, only manual crops (S0 vs S3.1 and S3.2) work slightly better.
In experiment S5, continuous and discrete pose estimation were compared. It was found out that regression works significantly worse. This tendency is also described in \cite{onoro2018challenge}, especially, when the amount of training images is not large enough.

\begin{figure}
\centering
\includegraphics[width=3.1in]{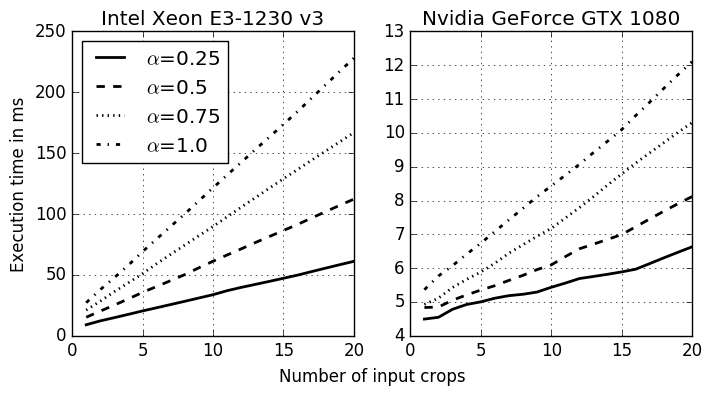}
\caption{Runtime analysis of the second stage for different MobileNet width multipliers $\alpha$ and number of robots (100~runs; standard deviation 45\% on CPU and 24\% on GPU).}
\label{fig_benchmark_second_stage}
\end{figure}

The inference time of the second stage is mainly affected by the width multiplier $\alpha$ and the number of robots detected by the first stage (see Figure \ref{fig_benchmark_second_stage}). For the default configuration ($\alpha = 0.5$) and an amount of 10 robots within the field of view of the camera, the second stage consumes 6~ms on a Nvidia GeForce GTX 1080. So both stages spend about 20~ms for one camera frame allowing a frame rate of 50~Hz, which is simliar to \cite{poirson2016single}.

To put it in a nutshell, we recommend the following settings to achieve a good second stage performance. Generating more images can increase the performance a lot. The network size should be adapted to the task complexity providing enough feature maps to tackle the problem but not too much to prevent overfitting. If possible, multiple background sources should be used. The crop method does not influence the performance significantly and can be chosen according to own preference.

\subsubsection{Total Performance}

\begin{figure}
\centering
\includegraphics[width=3.1in]{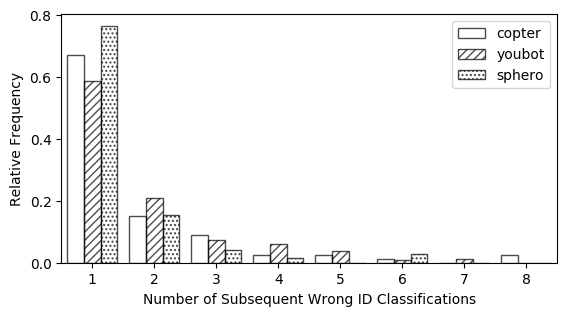}
\caption{Distribution of successive wrong robot type/instance detections.}
\label{fig_subsequent_wrong_classification_distribution}
\end{figure}

\begin{table}[b]
\caption{Performance of the whole framework}
\label{table_two_stage_performance}
\begin{center}
\begin{tabular}{|l||c|c|c|}
\hline
 & Copter & Sphero & YouBot\\
\hline \hline
mAP@0.5IOU & 97.9\% & 70.0\% & 96.6\%\\
\hline
Orientation MAE & 1.6$^{\circ}$ & 11.9$^{\circ}$ & 2.6$^{\circ}$\\
\hline
\end{tabular}
\end{center}
\end{table}

Finally, we have chosen the best performing configurations of the experiments of both stages (see Section \ref{first_stage} and \ref{second_stage}) for an integrated performance evaluation of the framework. For the first stage, the setting of F1.2 (trained only on COCO backgrounds) and, for the second stage, the setting of F2.2 (8,000 training images per identification) was applied. The framework was evaluated on complete video streams of the camera under application conditions with 10 Hz frame rate.

In Table \ref{table_two_stage_performance}, the mAP@0.5IOU for the robot type and instance detection of the whole framework as well as the mean absolute error of the orientation for correctly detected robots are shown. The good results of both stages are maintained in the framework. Depending on robot type, it achieves 70\% - 98\% detection mAP@0.5IOU and 2$^{\circ}$ - 12$^{\circ}$ orientation error.

Moreover, the number of successive wrong detections/classifications in a row was evaluated. The results are illustrated in Figure \ref{fig_subsequent_wrong_classification_distribution}. It can be seen that the framework misses the correct classification of a robot only a few frames in a row (usually less than five), which can easily be post-processed e.g. by an Extended Kalman filter to improve the performance even further.

\addtolength{\textheight}{-0.1cm}   

\FloatBarrier

\section{CONCLUSIONS AND FUTURE WORK}
\label{conclusion}

This paper shows our new adaptive robot tracking framework. The architecture
aims to enable an easy setup of a robot tracking system for arbitrary robots
using common camera hardware by exploiting machine learning. The delegation of
the configuration of the system to machine learning with CNNs majorly eases the
setup. The user only needs to provide appropriate training data in the form of
video streams acquired by the camera showing the robots to track as well as the
current orientation of the robots in the video stream. The system can then
automatically extract the necessary information and train the robot tracking
system. It was tested in a real laboratory with three types of robots with
different identification properties to show the adaptability. In general, the
performance of the system is very good and provides robot pose tracking
accuracy within a 0.5 intersection over union with less than $4\%$ of
identification error and an orientation error lower than $3^\circ$ for two of three
robot types. The higher orientation error of the third robot type is caused by
special physical properties of this type of robot and is not induced by the
system itself. The delay created by the tracking system was approximately
140~ms on typical PC hardware, but could be majorly decreased by exploiting
hardware acceleration through GPUs up to 20~ms. This provides the tracked
robots with localization information better than \emph{GPS}.

In our experiments, we have found out that the amount of generated images is a
major factor increasing the performance of the system. Moreover, it can be
trained using generic background images without the need for scenario-specific
backgrounds.

The current approach was evaluated using a single camera, but we aim to use
multiple cameras in the future to extend the detection area and avoid occlusion
of robots. Additionally, we want to evaluate different meta-architectures and CNNs
for the implementation of the first and second stage of the system such as
SSDLite to enhance the update rate.

We hope to minimize the
effort for the user even further. Currently, the system enforces the user to choose between
labeling effort and performance of the system. Even though the difference is
not big, we aim to improve the automatic crop to provide at least the same
output performance as the manual one. Additionally, we want to evaluate the
single stage approach to see if we can minimize the latency of the detection or
provide better detection, identification and pose estimation quality. Finally,
we intend to extend our compositor to provide better training data, especially
regarding perspective-awareness enabling more flexible camera positioning.




\bibliographystyle{IEEEtranS}
\bibliography{References}

\end{document}